\documentclass[10pt,twocolumn,letterpaper]{article}

\usepackage[utf8]{inputenc}
\usepackage{iccv}
\usepackage{times}
\usepackage{epsfig}
\usepackage{graphicx}
\usepackage{amsmath}
\usepackage{amssymb}

\usepackage[breaklinks=true,bookmarks=false]{hyperref}

\iccvfinalcopy %

\def\iccvPaperID{6294} %

\ificcvfinal\fi

\begin{document}

% !TEX root = ../top_final.tex
% !TEX spellcheck = en-US

%\definecolor{olive}{RGB}{50,150,50}

\newif\ifdraft
\drafttrue

\ifdraft
\newcommand{\PF}[1]{{\color{red}{\bf pf: #1}}}
\newcommand{\pf}[1]{{\color{red} #1}}
\newcommand{\HR}[1]{{\color{blue}{\bf hr: #1}}}
\newcommand{\hr}[1]{{\color{blue} #1}}
\newcommand{\VC}[1]{{\color{green}{\bf vc: #1}}}
\newcommand{\vc}[1]{{\color{green} #1}}
\newcommand{\ms}[1]{{\color{olive}{#1}}}
\newcommand{\MS}[1]{{\color{olive}{\bf ms: #1}}}
\newcommand{\JS}[1]{{\color{cyan}{\bf js: #1}}}
\newcommand{\NEW}[1]{{\color{red}{#1}}}

\else
\newcommand{\PF}[1]{{\color{red}{}}}	
\newcommand{\pf}[1]{ #1 }
\newcommand{\HR}[1]{{\color{blue}{}}}
\newcommand{\hr}[1]{#1}%
\newcommand{\VC}[1]{{\color{green}{}}}
\newcommand{\ms}[1]{ #1 }
\newcommand{\MS}[1]{{\color{green}{}}}
\newcommand{\NEW}[1]{#1}
\fi

%%% unit definitions %%%
\newcommand{\um}{\text{m}} % meters
\newcommand{\us}{\text{s}} % seconds
\newcommand{\upx}{\text{px}} % seconds

%%% Vector definitions %%%
\newcommand{\va}{\mathbf{a}}
\newcommand{\vb}{\mathbf{b}}
\newcommand{\vd}{\mathbf{d}}
\newcommand{\ve}{\mathbf{e}}
\newcommand{\vf}{\mathbf{f}}
\newcommand{\vg}{\mathbf{g}}
\newcommand{\vh}{\mathbf{h}}
\newcommand{\vi}{\mathbf{i}}
\newcommand{\vj}{\mathbf{j}}
\newcommand{\vk}{\mathbf{k}}
\newcommand{\vl}{\mathbf{l}}
\newcommand{\vm}{\mathbf{m}}
\newcommand{\vn}{\mathbf{n}}
\newcommand{\vo}{\mathbf{o}}
\newcommand{\vp}{\mathbf{p}}
\newcommand{\vq}{\mathbf{q}}
\newcommand{\vr}{\mathbf{r}}
\renewcommand{\vs}{\mathbf{s}} % note, renw instead of new as already defined elsewhere
\newcommand{\vt}{\mathbf{t}}
\newcommand{\vu}{\mathbf{u}}
\newcommand{\vv}{\mathbf{v}}
\newcommand{\vw}{\mathbf{w}}
\newcommand{\vx}{\mathbf{x}}
\newcommand{\vy}{\mathbf{y}}
\newcommand{\vz}{\mathbf{z}}

%%% matrix definitions %%%
\newcommand{\mA}{\mathbf{A}}
\newcommand{\mB}{\mathbf{B}}
\newcommand{\mC}{\mathbf{C}}
\newcommand{\mD}{\mathbf{D}}
\newcommand{\mE}{\mathbf{E}}
\newcommand{\mF}{\mathbf{F}}
\newcommand{\mG}{\mathbf{G}}
\newcommand{\mH}{\mathbf{H}}
\newcommand{\mI}{\mathbf{I}}
\newcommand{\mJ}{\mathbf{J}}
\newcommand{\mK}{\mathbf{K}}
\newcommand{\mL}{\mathbf{L}}
\newcommand{\mM}{\mathbf{M}}
\newcommand{\mN}{\mathbf{N}}
\newcommand{\mO}{\mathbf{O}}
\newcommand{\mP}{\mathbf{P}}
\newcommand{\mQ}{\mathbf{Q}}
\newcommand{\mR}{\mathbf{R}}
\newcommand{\mS}{\mathbf{S}}
\newcommand{\mT}{\mathbf{T}}
\newcommand{\mU}{\mathbf{U}}
\newcommand{\mV}{\mathbf{V}}
\newcommand{\mW}{\mathbf{W}}
\newcommand{\mX}{\mathbf{X}}
\newcommand{\mY}{\mathbf{Y}}
\newcommand{\mZ}{\mathbf{Z}}

\newcommand{\cA}{\mathcal A}
\newcommand{\cB}{\mathcal B}
\newcommand{\cC}{\mathcal C}
\newcommand{\cD}{\mathcal D}
\newcommand{\cE}{\mathcal E}
\newcommand{\cF}{\mathcal F}
\newcommand{\cG}{\mathcal G}
\newcommand{\cH}{\mathcal H}
\newcommand{\cI}{\mathcal I}
\newcommand{\cJ}{\mathcal J}
\newcommand{\cK}{\mathcal K}
\newcommand{\cL}{\mathcal L}
\newcommand{\cM}{\mathcal M}
\newcommand{\cN}{\mathcal N}
\newcommand{\cO}{\mathcal O}
\newcommand{\cP}{\mathcal P}
\newcommand{\cQ}{\mathcal Q}
\newcommand{\cR}{\mathcal R}
\newcommand{\cS}{\mathcal S}
\newcommand{\cT}{\mathcal T}
\newcommand{\cU}{\mathcal U}
\newcommand{\cV}{\mathcal V}
\newcommand{\cW}{\mathcal W}
\newcommand{\cX}{\mathcal X}
\newcommand{\cY}{\mathcal Y}
\newcommand{\cZ}{\mathcal Z}

\newcommand{\TODO}[1]{\textcolor{cyan}{#1}}

\newcommand{\ST}{\mathcal{T}}
\newcommand{\SST}{\mathcal{T}_S}

\newcommand{\R}{\mathbb{R}}
\newcommand{\Seg}{\mathbf{S}} % geometric part
\newcommand{\Latent}{\mathbf{L}}
\newcommand{\LatentG}{\Latent^{\text{3D}}} % geometric part
\newcommand{\LatentA}{\Latent^\text{app}} % appearance part
\newcommand{\LatentBG}{\mB} % appearance part

\newcommand{\comment}[1]{}

\newcommand{\norm}[1]{\left\lVert#1\right\rVert}
\newcommand{\argmin}{\operatornamewithlimits{argmin}}
\newcommand{\erf}{\operatornamewithlimits{erf}}

\newcommand{\parag}[1]{\vspace{-3mm}\paragraph{#1}}

\newcommand{\curve}[0]{{\bf Ours}} %-curve
\newcommand{\dist}[0]{{\bf Baseline-distance-based}}
\newcommand{\population}[0]{{\bf Baseline-population-mean}}

\newcommand{\ball}[0]{{\bf FallingBall}}
\newcommand{\persons}[0]{{\bf ArticulatedFreeFall}}

\title{Gravity as a Reference for Estimating a Person's Height from Video}

\author{
Didier Bieler$^1$
\and
Semih Günel$^1$
\and
Pascal Fua$^1$
\and
Helge Rhodin$^{1,2}$
\and
$^1$EPFL, Lausanne, Switzerland\quad
$^2$UBC, Vancouver, Canada\\
{\tt\small firstname.lastname@epfl.ch}
}
\maketitle
\ificcvfinal\thispagestyle{empty}\fi

\begin{abstract}
Estimating the metric height of a person from monocular imagery without additional assumptions is ill-posed. Existing solutions either require manual calibration of ground plane and camera geometry, special cameras, or reference objects of known size. 
We focus on motion cues and exploit gravity on earth as an omnipresent reference 'object' to translate acceleration, and subsequently height, measured in image-pixels to values in meters.
We require videos of motion as input, where gravity is the only external force. This limitation is different to those of existing solutions that recover a person's height and, therefore, our method opens up new application fields.  
We show theoretically and empirically that a simple motion trajectory analysis suffices to translate from pixel measurements to the person's metric height, reaching a MAE of up to 3.9 cm on jumping motions, and that this works without camera and ground plane calibration.  
\end{abstract}

\section{Introduction}

Estimating metric scale from a monocular image or video recordings is a fundamental problem in computer vision \cite{Criminisi00,Hartley00} and important for determining distances in forensics, autonomous driving, person re-identification, and structure-from-motion (SfM).
In general, object size and distance cancel in perspective projection---which makes the problem ill-posed.
However, solutions exist if cameras and ground floor are manually calibrated \cite{Guan09,Zhou16d,Li11d}, special cameras for depth-of-field sweeping are available \cite{Georgiev13}, or reference objects of known scale are present \cite{Vester12,Hartley00,Vester12,Ljungberg08}. Some solutions studied relationships that are particular for persons, such as height, appearance, facial features, body proportions \cite{Benabdelkader08,Gunel18}. However, only uncertain and biased predictions could be obtained.

\begin{figure}[t]
	\centering
	\vspace{-0.7cm}
	\includegraphics[height=5cm]{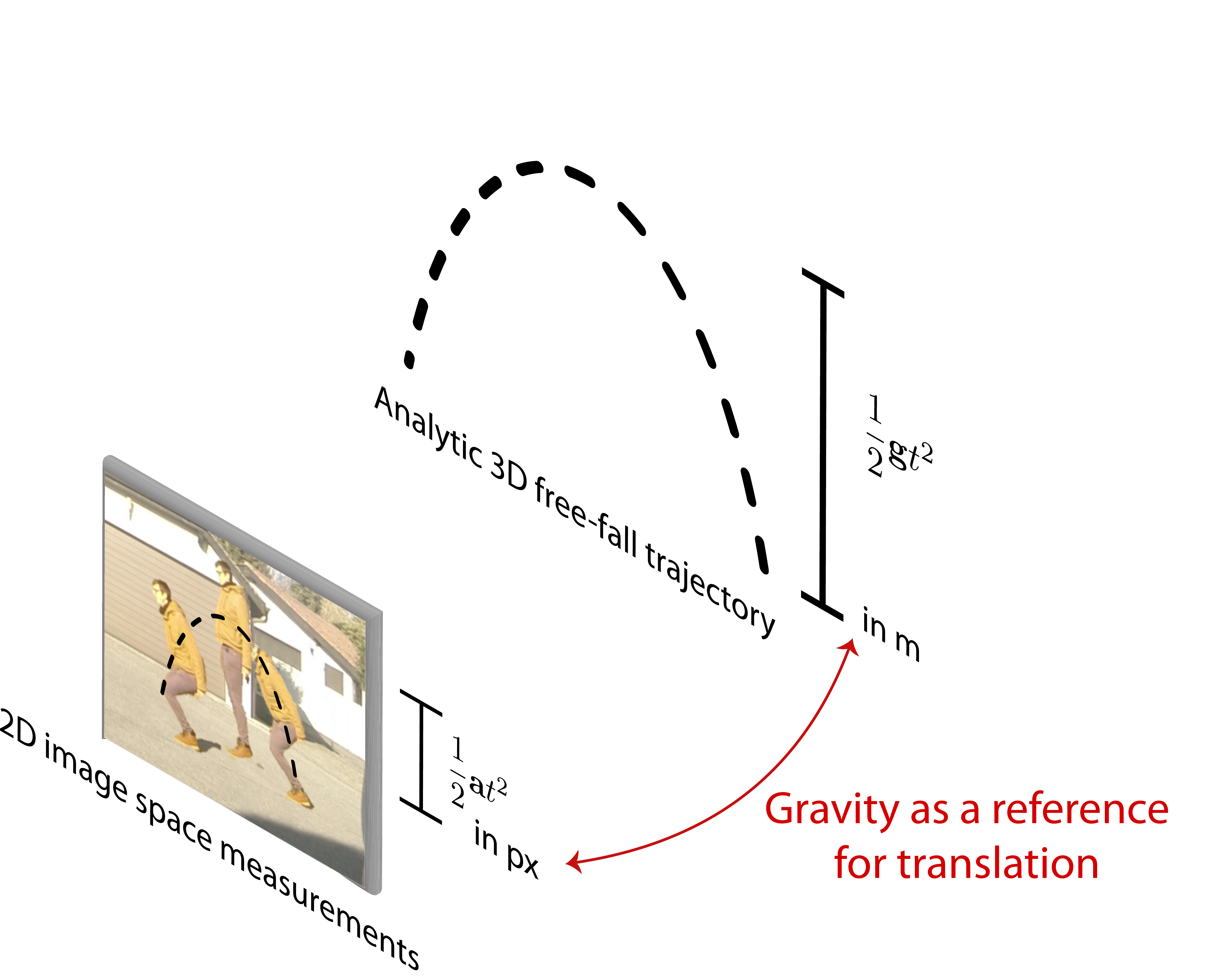}
	\caption{\textbf{Idea.} We exploit gravity as a reference \emph{object} for mapping image height measurements in pixel [px] to metric height [m].}
	\label{fig:teaser}
\end{figure}

This paper aims at a new approach to estimating a person's height using motion cues in videos.
The main idea is to use the omnipresent gravity on earth as a reference 'object'. Newton's second equation of motion dictates that the trajectory of an object is a parabola, a function of time, its initial speed and position, with the curvature determined by the acceleration induced by constant external forces \cite{Newton1729}. 
To this end, we restrict ourselves to cases where gravitation is the only source of external acceleration and the camera is static, so that acceleration in the image can be uniquely attributed to gravity.
By relating acceleration in the image and to gravity on earth, we can then translate measurements in pixels to metric height in meters, as sketched in Figure~\ref{fig:teaser}.

\begin{figure}[t]
	\centering
	\includegraphics[width=\linewidth]{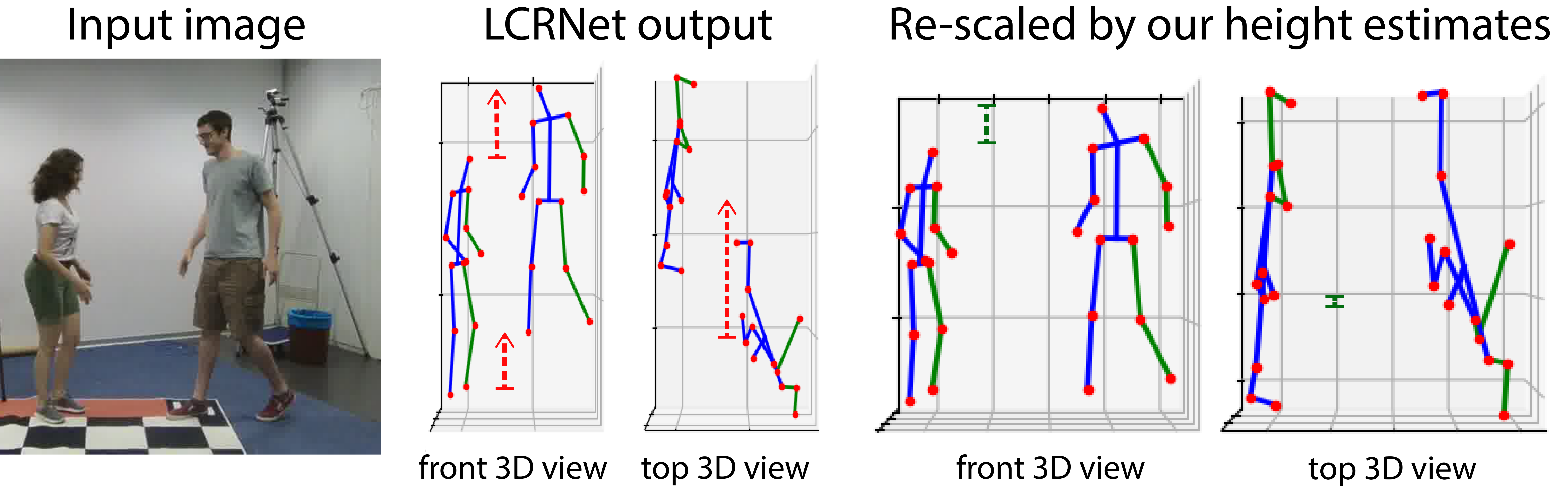}
	\caption{\textbf{Application to metric 3D pose estimation. Left:} 
		Input and output of LCRNet, a 3D human pose estimation method~\cite{Rogez18}, which does not recover the correct scales (marked in red).
		\textbf{Right:} Using our height estimate from directed on-spot jumping recovers the scale and relative depth of the hip locations (marked in green).}
	\label{fig:metricPose}
\end{figure}

Although this strategy restricts the application to scenarios entailing free-fall motions, it applies to any video of a person jumping or running, where air friction and other sources of acceleration are negligible. 
Because this limitation is orthogonal to those of existing solutions, our method opens up new application fields, such as metric monocular SfM, automatic person re-identification in videos with unknown camera geometry, and metric human pose estimation from a single camera, as demonstrated in Figure~\ref{fig:metricPose}. 

Our method is inspired by approaches that estimate the 3D trajectory of rigid objects in free fall \cite{kim98,Ohno00,Ribnick09,Kumar11,Skold15}. 
All these methods assume a calibrated camera, known gravity direction, rigid objects and focus on object position instead of scale estimation.
In contrast, we show that our strategy for height estimation is applicable without knowing or constraining the initial object speed or position, that it can be generalized to account for the internal forces present in articulated person motion, and does not require knowledge of any camera parameters nor ground plane position.

We validate our findings on a new test set containing 12 persons performing seven different jumping and running motions at two distances. Our obtained mean absolute error (MAE) of 3.9 cm, is nearly half of the 6.5 cm MAE by \cite{Gunel18}, which demonstrates the success our our method and the importance of geometric and physical constraints. %
Our code and dataset is available at \url{https://cvlab.epfl.ch/articulated-free-fall-dataset}.

\section{Related Work}
\label{sec:related}

Multiple approaches for estimating the size of a person or object have been proposed in the literature. We list the most important ones, focusing on the existing conditions used to make monocular height estimation well-posed.

\paragraph{Height from camera geometry.}
Without external scale information, object size and distance is ambiguous according to the
basic pinhole camera model. In practice, lenses have a limited depth of field,
which shape-from-defocus techniques exploit~\cite{Mather96,Shi15b}.
While it can be used to guess depth orderings in a single image, a
focal sweep across multiple images or a specialized camera~\cite{Georgiev13} is
required for metric scale reconstruction.
These constraints preclude applications to monocular height estimation from single images and videos recorded with consumer cameras.

\paragraph{Scale from scene and camera geometry.}
One method for estimating height from images it to place several reference objects nearby the target object \cite{Vester12,Ljungberg08}. Placing multiple references is cumbersome and has been overcome 
by relating a single reference to other points in the scene through vanishing points of the ground plane~\cite{Criminisi00,Hartley00,Vester12}.
An alternative is to manually calibrate the camera height and orientation in
relation to the ground plane~\cite{Guan09,Zhou16d,Li11d}.
Thereby, the height of standing people can be inferred by locating the head and foot position in a single image, triangulating the 3D foot position on the known ground plane, and scaling the head-foot distance to this 3D position.
These methods do not require to alter the scene but they are not applicable to legacy videos and require an expert to calibrate.

\paragraph{Height from statistics.}

Dey et al.~\cite{Dey14} propose a unique solution that exploits statistics across a group of people by measuring relative heights in group pictures and connecting these in large image collections.
Absolute height is then estimated from the resulting network of relative heights by enforcing consistency with the average human. %
Medical studies determined a correlation of human height and ratios of limb proportions~\cite{Adjeroh10} and the ratio of head to shoulders \cite{Shiang99,Kato98}, but these are difficult to estimate from images.
Recent methods have attempted to capture such statistical relations directly from anthropometric
measurements~\cite{Benabdelkader08,Gordon89} and collection of images \cite{Gunel18,Dantcheva18} though black-box regression using linear regression and deep learning, respectively.
We show that these statistical methods suffer from errors and bias towards the average human height, as monocular scale estimation is ill-posed without geometric constraints.

\paragraph{Physics-based trajectory modeling.}
Physical constraints are widely used for refining and estimating trajectories of rigid objects. 
Kim et al.~\cite{kim98} recover the 3D positions of a soccer ball in relation to the known height of a player or goalpost by exploiting that the ball trajectory follows a parabola in free flight. 
Ohno et al.~\cite{Ohno00} model gravity and air drag explicitly to recover metric 3D football trajectories from a calibrated camera. This strategy has also been shown to generalize to other projectiles \cite{Ribnick09}.
Kumar et al. \cite{Kumar11} analyze tennis and use physics to fill-in frames for which no multi–view triangulation of the ball is available.
All of these methods assume a calibrated camera, rigid object, focus on object position, and, thereby, do not address our aim of recovering a person's height from uncalibrated video.

\section{Theory}
\label{sec:theory}

Gravitation is omnipresent and roughly constant on the earth's surface, with an acceleration of $g \approx 9.81 {m}/{s^2}$ towards the center of the planet and the variation across the surface below four percent. 
\footnote{Extremes are, e.g., $9.78 {m}/{s^2}$ in Singapore and $9.83 {m}/{s^2}$ in Oslo.}

A rigid object in free-fall, where gravitation is the only acceleration, is explained by Newton's equation of motion:
\begin{equation}
\vp(t) = \frac{1}{2} \vg t^2 + \vv_0 t + \vp_0,
\label{eq:motion}
\end{equation}
where $\vp_0$ and $\vv_0$ are 3D-vectors capturing the initial position and velocity, respectively, $t$ the elapsed time, and $\vg = \vn g$ the acceleration in direction $\vn$ (vertically, depending on the chosen coordinate system). 

The underlying idea is to measure motion in video and relate the quantities estimated in pixel units to metric meters through Eq.~\ref{eq:motion}. Figure~\ref{fig:teaser} sketches this relation graphically. 
For example, in the special case of $\vv_0=0$, Newton's equation dictates that an object moves $4.9 m$ after one second of free fall. Hence, estimating a motion of $|\vp^\text{px}|$ pixels (px) over a video of length $t=1\us$ yields $1 \upx = \frac{1}{|\vp^\text{px}|} 4.9 \um$, uniquely determining the px to meter ratio, which in turn allows to translate height measurements of the moving object in the image to metric units in 3D world coordinates. Next, we derive this relation formally with respect to the camera position and orientation. We first derive them for the case of rigid objects and then for articulated motion.

\subsection{Computing a Rigid Object's Height}

In this section we derive a linear relation, the factor $q$, that maps the distances measured in the image, e.g., height $h^\text{px}$ measured from upper to lower object extend, to corresponding height in meters,
$h = h^\text{px} q$.
 Notably, this factor can be computed for unknown distance $d$, focal length $f$, and gravity direction, by relating the gravitational constant $g$ to the measured image acceleration $\va^\text{px}$.

Formally, the observed 2D motion $(\vp^\text{px}_1,\dots,\vp^\text{px}_T)$ on the image plane across a video of $T$ frames are samples of the projected 3D motion $\vp^\text{px}(t) = \Pi\left( \vp(t)\right)$. In this work, we will approximate the projection process with scaled orthographic projection, $\Pi$. In terms of camera coordinates, with optical axis pointing towards the third coordinate, the projected free-fall motion (Eq.~\ref{eq:motion}) is then explained by
\begin{align}
\vp^\text{px}(t) &= \frac{f}{d} \begin{bmatrix}
1 & 0 & 0\\
0 & 1 & 0
\end{bmatrix}\vp(t) \nonumber\\
&= \frac{f}{d} \begin{bmatrix}
1 & 0 & 0\\
0 & 1 & 0
\end{bmatrix} \left(\frac{t^2}{2} \vg + \vv_0 {t} + \vp_0 \right),
\label{eq:projected motion}
\end{align}
here $d$ is the object distance to the camera. This projection formula implies that the image motion is also a parabola, 
\begin{equation}
{\vp}^\text{px}(t) = \frac{1}{2} \va^\text{px} t^2 + \vv^\text{px}_0 t + \vp^\text{px}_0 \;.
\label{eq:image motion}
\end{equation}
The estimation of $\va^\text{px}, \vv^\text{px}_0$ and $\vp^\text{px}_0$ from the input video is explained in Section~\ref{sec:acceleration}.
By relating the quadratic terms in Eq.~\ref{eq:projected motion} and Eq.~\ref{eq:image motion} %
, we obtain the following relation between the measured 2D and predicted 3D acceleration,
\begin{align}
\frac{1}{2} \va^\text{px} t^2
=& \frac{f}{d} \begin{bmatrix}
1 & 0 & 0\\
0 & 1 & 0
\end{bmatrix} \frac{t^2}{2} \vg
\nonumber \\
\Leftrightarrow \quad\quad
\va^\text{px}
=& \frac{f}{d} \begin{bmatrix}
1 & 0 & 0\\
0 & 1 & 0
\end{bmatrix} \vn g.
\label{eq:projection}
\end{align}
This relation is sketched in Fig.~\ref{fig:teaser}. Although $\vn, f$, and $d$ are unknown, they are constants. For an object at $d$, their combined effect is determined by the acceleration quotient, 
\begin{align}
\vq = \frac{d}{f \vn^\text{px}} = \frac{g}{\va^\text{px}}
\quad\text{ with }\quad
\vn^\text{px} = \begin{bmatrix}
1 & 0 & 0\\
0 & 1 & 0
\end{bmatrix} \vn,
\label{eq:translation}
\end{align}
where $g$ is known, $\va^\text{px}$ is the observed acceleration in Eq.~\ref{eq:image motion}. Note that, vector-scalar operations are here element-wise.

\begin{figure}[t]
	\centering
	\includegraphics[width=0.7\columnwidth]{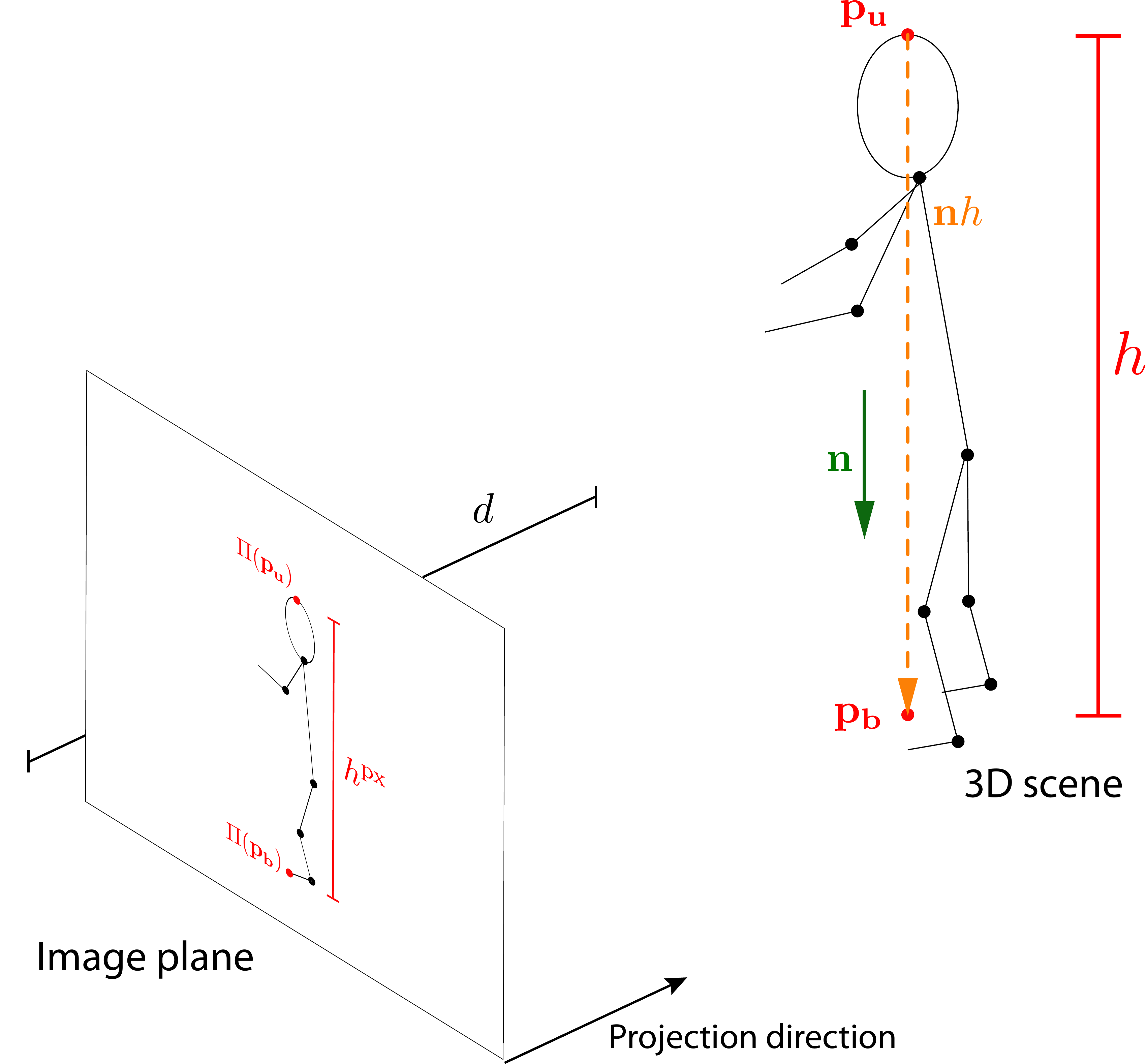}
	\caption{\textbf{Projection of height.} The projected height, $h^\text{px}$ is the distance between two projected points $\Pi(\vp_u)$ and $\Pi(\vp_b)$ that are aligned with the direction of gravity $\vn$ in 3D and span height $h$.}
	\label{fig:height projection}
\end{figure}

We now turn to the height estimation using $\vq$. We define height as the distance between two 3D points $\vp_u$ and $\vp_b$ that are in a line with the direction of gravity $\vn$, see Figure~\ref{fig:height projection}. The sought function is derived by applying $\Pi$ on $\vp_u$ and $\vp_b$, using the linearity of $\Pi$, and substituting $\vq$ from Eq.~\ref{eq:translation}, 
\begin{align}
\vh^\text{px} &= \Pi(\vp_u)-\Pi(\vp_b)  \nonumber\\
&= \frac{f}{d} \begin{bmatrix}
1 & 0 & 0\\
0 & 1 & 0
\end{bmatrix} \left(\vp_u- \vp_b \right) \nonumber\\
&= \frac{f}{d} \begin{bmatrix}
1 & 0 & 0\\
0 & 1 & 0
\end{bmatrix} \vn h 
= \frac{h}{\vq}  
\;, \label{eq:height projection}
\end{align}
where, $\vh^\text{px}$ is the difference between the projected reference points used for height estimation---in our experiments it is the head-to-heel vector measured in the image.
This relation provides two solutions for $h$, respectively for the vertical and horizontal components of $\vq$ and $\va^\text{px}$. This system of equations could be solved in the least-squares sense. 

However, in practice, videos are predominantly captured upright and the vertical direction will dominante. We therefore use the vertical solution directly. In the following, we denote the vertical components of $\vq, \va^\text{px}$ and $\vh^\text{px}$ respectively with scalars $q,a^\text{px}$ and $h^\text{px}$. 
The translation rule between $h^\text{px}$ in $\upx$ and metric height is~then
\begin{align}
h = h^\text{px} q
\;.\label{eq:height translation}
\end{align}

Note that the gravity direction and camera projection matrix are subsumed in $q$. Hence, equations hold for any affine camera model and no ground plane nor camera calibration is needed. Also the simplified Eq.\ref{eq:height translation} applies to unknown camera orientation, so long vertical acceleration is non-zero. %

Only if $n$ and $f$ are known, i.e., when the direction of gravity, camera intrinsic, and extrinsic parameters are calibrated, it is true that $q$ can be further decomposed to compute the object's distance $d$ and extend in all directions, which was the focus of previous studies \cite{kim98,Ohno00,Ribnick09,Kumar11,Skold15}.

\subsection{Computing a Person's Height }

A jumping or otherwise moving person is likely to articulate the arms and other body parts actively. Instead of trying to model the complex interactions between body parts, we model the human body as a closed system and switch to Center of Mass (COM) computations. Conveniently, by the law of conservation of momentum, the center of mass of a closed system will move at a constant speed even if internal parts move in other ways, and it also follows
Newton's equation of motion (Eq.~\ref{eq:motion}) when accelerated by $\vg$. Moreover, the (scaled orthographic) projection of the 3D-COM is the 2D-COM of the individually projected body parts, so that Eq.~\ref{eq:projection} holds as well.

To sum up, to estimate the height of a person we propose to compute the COM trajectory in 2D, solve Eq.~\ref{eq:translation} for $q$, and apply $q$ in Eq.~\ref{eq:height translation} to translate the height $h^\text{px}$ from head to heel measured in pixel units to absolute height $h$ in m. 

In the following, we will explain how the needed quantities, such as COM trajectory can be inferred automatically from videos without requiring camera calibration.

\section{Algorithms}
\label{sec:method}

The algorithm splits into four consecutive steps: computing object position and COM per frame, detecting free fall events, estimating image acceleration, and measuring height in the image for total height computation.

\subsection{Measuring the COM}
\label{sec:COM}

For algorithmic validation, we analyze the ideal projectile trajectory of a rigid and uniformly colored ball. In this toy example, the image position measurements can easily be automated. We segment the ball from the background through color thresholding and determine the object center, $\vp^\text{px}(t)$, by fitting a circle to the segmentation contour.

\begin{figure}[t]
	\centering
	\includegraphics[width=\columnwidth]{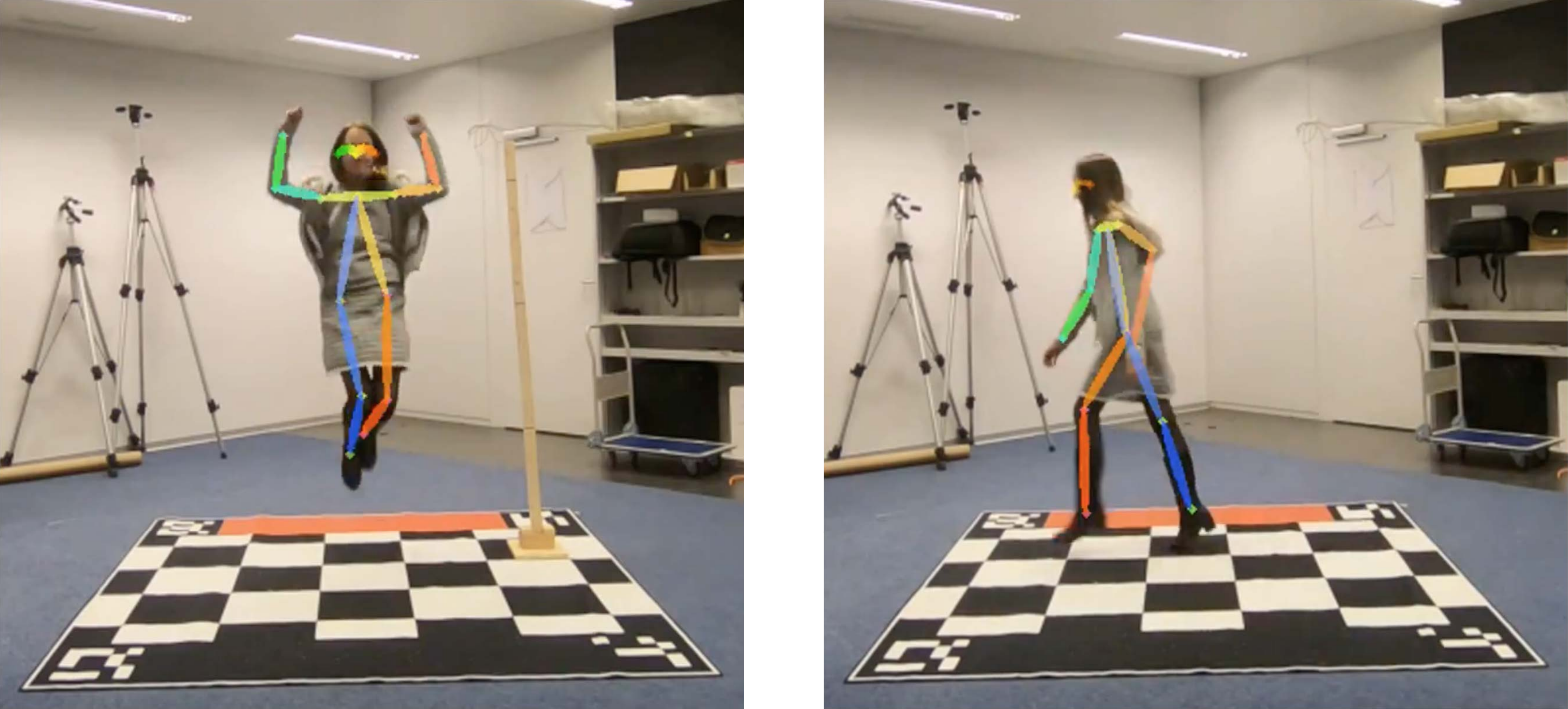}
	\caption{\textbf{Bodypart keypoint detections} using AlphaPose on frontal and lateral motion, displayed as colored skeleton overlay.}
	\label{fig:AlphaPose}
\end{figure}

For articulated human motion, we detect the person's body parts with AlphaPose \cite{Fang17b}, a neural network that has been trained to predict person keypoints in the input image.
Given a set of $T$ RGB images $I \in \R^{3\times W \times H}$ representing the $T$ frames of the original video,
AlphaPose outputs $J=17$ person 2D joint positions $(\vp^{\text{px},j}_t)_{j=1}^J$ for each frame $t$. Examples are shown in Fig.~\ref{fig:AlphaPose}. If multiple persons are detected, we select the largest one, the one in the foreground.

To compute the needed COM position, $\vp^\text{px}_t = \sum_{j=0}^{J} r_j \vp^{\text{px},j}_t$, of the $J$ parts at positions $\vp^{\text{px},j}_t$, one needs to know the ratio $r_j$ between the weight of body part $j$ in relation to the total weight.
For persons, the absolute weight varies significantly and it appears to be at least as hard to estimate weight as to recover the sought height.
However, the relative weight of bodyparts varies little across individuals. We took the mean mass distribution estimated in a large-scale study in \cite{Clauser71}, which has been widely used for COM computation in the past.

\subsection{Detecting Free-Fall}
\label{sec:free-fall}

\begin{figure}[t]
	\centering
	\includegraphics[width=8cm]{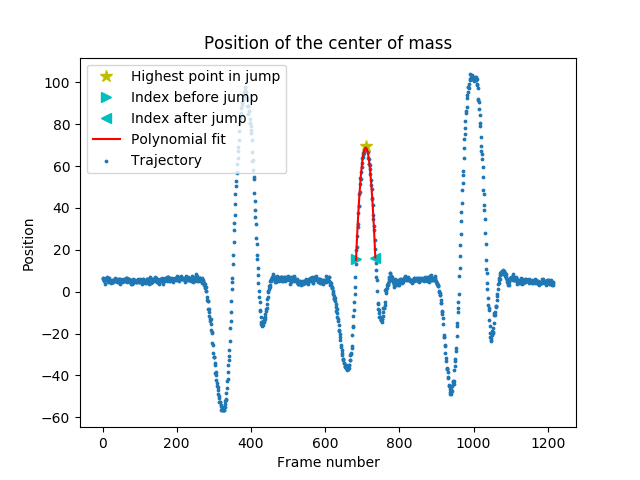}
	\caption{\textbf{COM trajectory for on-spot jumping}, with the polynomial fit,detected start, maximum, and end points marked.}
	\label{fig:trajectory}
\end{figure}

We assume that gravity is the only external force applied on the person (neglecting air friction). However, this is only true during jumps and other free-fall phases. Thus, we derive a simple yet effective algorithm for distinguishing flight and ground contact phases. First, we localize maxima in the trajectory, the highest points $M = \{m | p^\text{px}_m > p^\text{px}_t \text{ for } t \in [m-10,\dots,m+10]\}$ in the trajectory of vertical pixel locations $p^\text{px}_t$. Second, we compute the position of contact, $p^\text{px}_\text{floor}$ as the median across the first 100 frames, expecting that the motion starts in a standing pose. We then select the frame interval $ S_m = [\vp^\text{px}_s,\dots,\vp^\text{px}_e]$ such that all $\vp^\text{px}_t \in S_m$ are at least 15 \% above the ground in relation to the jump peak~$p^\text{px}_m$. This procedure ensures the points belonging to the jump initiation are excluded, those where the body is accelerated and in contact with the ground. A selection example is visualized in Figure~\ref{fig:trajectory}. 

For lateral motion the ground contact point varies dependent on the position in the image, lense distortion, and camera orientation. For these, we switch to an alternative approach and determine the distance between a maxima and the two neighboring minimum values and select only those points that are in the upper half.

\subsection{Estimating Acceleration}
\label{sec:acceleration}

A key challenge is the estimation of $\va^\text{px}$ from the COM trajectory. Without loss of generality, we assume that the camera roll is small and, hence, acceleration is predominantly vertical. This can be ensured by rotating the image in the direction of maximal acceleration.
Moreover, we read the camera frame rate from the video meta data, to measure acceleration in pixel per seconds instead of number of frames. We propose two ways of estimating the image acceleration:

\parag{\dist{}.} At the highest point of a jump the velocity is zero. Hence, the linear term in Eq.~\ref{eq:motion} must be zero. Therefore, motion in the video after this turning point till inception with the ground can be uniquely attributed to gravitational acceleration. Taking the highest point $p_m$ and the last point before ground contact, $p^\text{px}_e$, we can easily solve $p^\text{px}_e - p^\text{px}_m = \frac{1}{2} a^\text{px} (e-m)^2$ for $a^\text{px}$ since $\vv_0^\text{px}$ is zero and $p^\text{px}_0$ constant. Here $(e-m)$ is the time difference in seconds. However, these point estimates are prone to error and the highest point might happen to be in-between two samples.%

\parag{\curve{}.} Under the free-fall assumption, the COM trajectory must follow a second order polynomial, with the quadratic term representing the acceleration and the linear term the velocity in pixels. Therefore, we fit a polynomial of degree two on the curve in the least squares sense. 
Since the COM trajectory is estimated through AlphaPose, wrong and inaccurate detections occur. We test two measures to counteract. First, we utilize the confidence output of AlphaPose and exclude points where the score drops below 2.
Second, we apply random sample consensus (RANSAC) \cite{Fischler81} on top. 

\begin{figure}[t!]
	\centering
	\includegraphics[width=7.5cm]{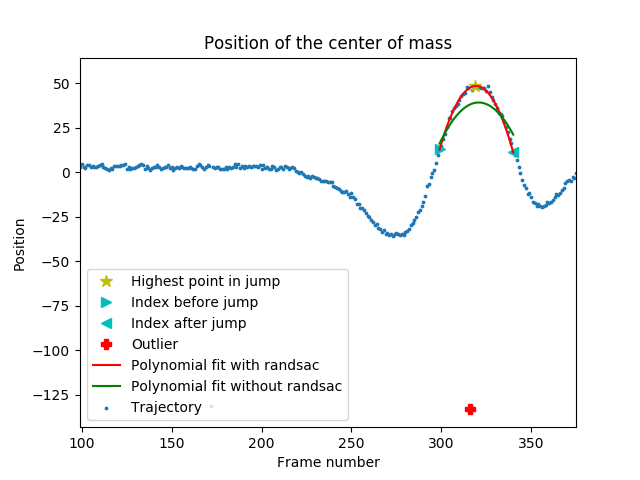}
	\caption{\textbf{Outlier removal example.} The red diamond marks the outlier and the red curve is the one that excludes it during fitting, leading to significantly improved fits.}
	\label{fig:useRandsac}
\end{figure}

The fitting process and the effect of outliers rejection is exemplified in Figure~\ref{fig:useRandsac}. 
The acceleration $\va^\text{px}$ is then simply two-times the quadratic coefficient of the curve fit. Note that the image acceleration needs to be measured in pixels per seconds squared using the video frame rate information.

\subsection{Converting Length Estimates to Human Height}
\label{sec:translation}

To estimate the total height of a person, we need to know the pixel location of the top of the head to the heel in a standing pose. Unfortunately, no such off-the-shelf detectors exist. The most utmost locations that AlphaPose returns are the nose and ankle points. As a stop-gap solution, we propose to infer a correction factor to go from nose-to-ankle to total height. 
We compute the mean ratio, $c=1.17\pm 0.03$, between the person's pixel height and ankle-nose distance, determined by AlphaPose, over 29 images taken from the web. The low standard deviation suggests that this linear approximation is accurate.

In practice, we measure the pixel height at the first frames of the video, assuming an upright stance. The final height is subsequently recovered by translating that image height measurements to meters with Eq.~\ref{eq:height translation} and multiplying the result with $c$.

\section{Experimental Evaluation}
\label{sec:eval}

\begin{figure}[t]
	\begin{center}
		\includegraphics[width=0.8\linewidth]{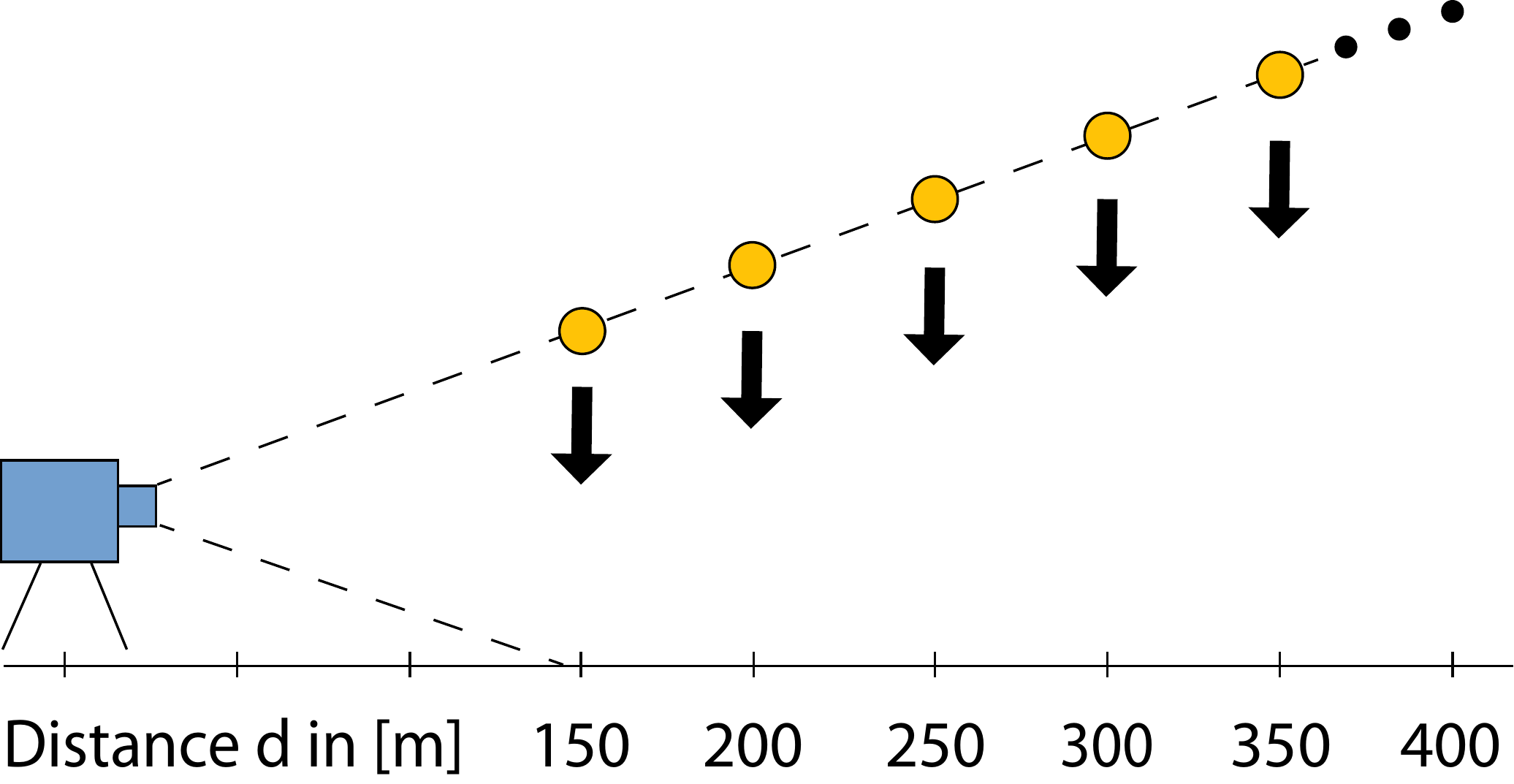}
		\caption{\textbf{Illustration of the tennisball experiment.} The ball is dropped in front of the camera so that it is still in the field of view.}
		\label{fig:tennis}
	\end{center}
\end{figure}

\begin{figure}[t]
	\begin{center}
		\includegraphics[width=\linewidth]{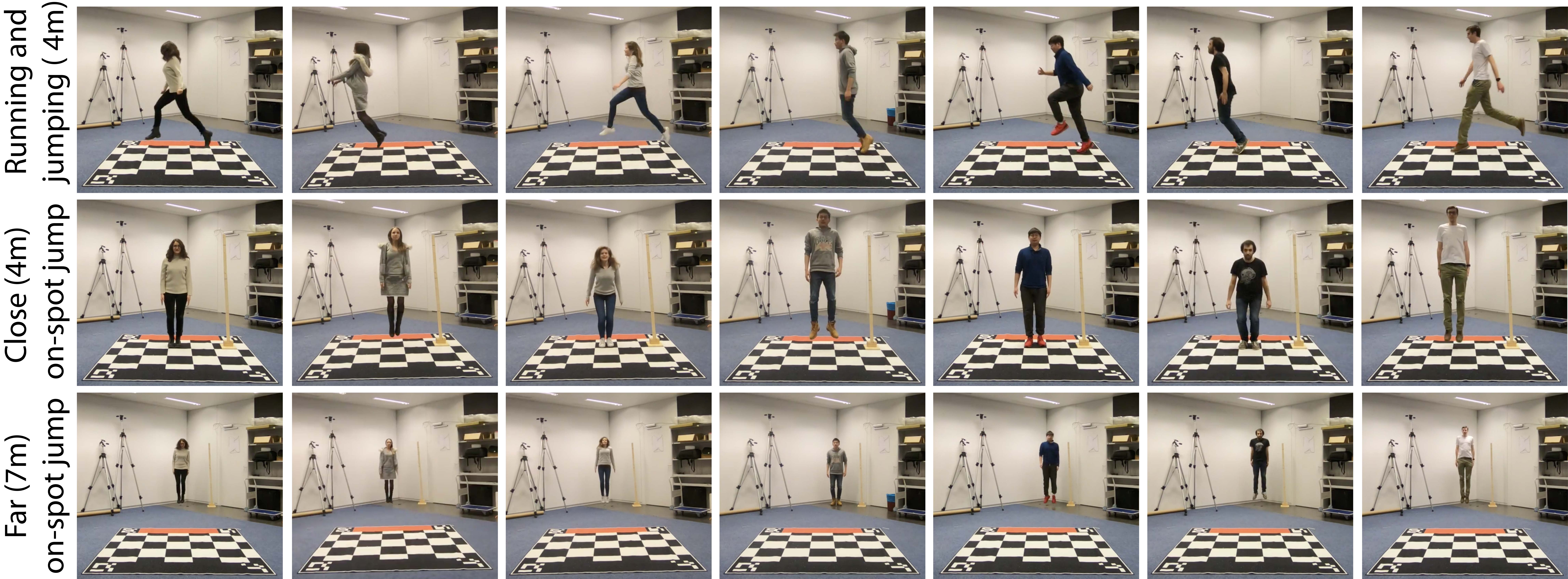}
		\caption{\textbf{Examples from our new \persons{} dataset}, that is captured at 4 and 7 m distance s and features lateral running, jumping and on-spot jumping.}
		\label{fig:dataset people}
	\end{center}
\end{figure}

\begin{figure}[t]
	\centering
	\includegraphics[width=\linewidth]{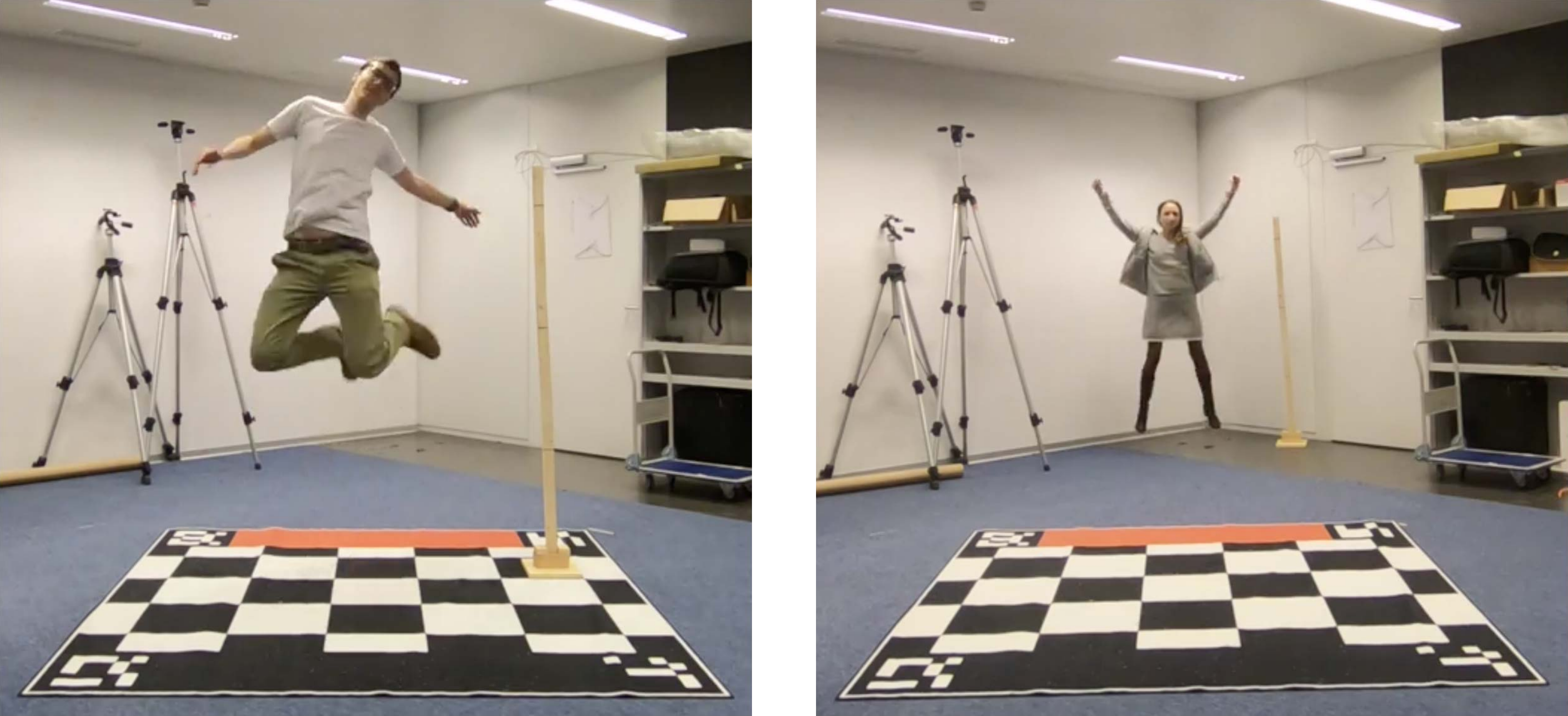}
	\caption{\textbf{Example of funny and jumping jack poses.} The wooden reference object is used to estimate relative errors of $q$.}
	\label{fig:funny}
\end{figure}

We first study the attainable accuracy of the gravity-based height estimation at hand of rigid objects, where automatic detection is easy and reliable.
Subsequently, we analyze the feasibility of estimating a person's height during jumps and runs of varying complexity and compare it to ground truth measurements and to the appearance-based solution proposed in \cite{Gunel18}.
We introduce new tests sets for both setups:
\begin{itemize}
	\item \ball{}. We drop a tennisball at several distances to the camera, spaced 50cm apart, as sketched in Figure~\ref{fig:tennis}. Two bounces of the ball are recorded at 120 Hz, the second one is about 50\% lower due to the absorbed energy at inception with the ground. The introduced ball detection algorithm is used to calculate the ball diameter in pixels. The ground truth diameter is 7.3 cm.
	\item \persons{}. We recorded {12} subjects, at 30 Hz, located at {four} and {seven} meter distance to the camera. Examples are shown in Figure~\ref{fig:dataset people}. {Seven} motions are tested: low, high, jumping jack, and funny on-spot jumps, as well as lateral running, exaggerated running, and jumping with swinging arms. Funny jumps are undirected and participants chose asymmetric, articulated poses, see Figure~\ref{fig:funny} for examples. The dataset is available at \url{https://cvlab.epfl.ch/articulated-free-fall-dataset}.
\end{itemize}

On top of this, we test human height estimation on community videos downloaded from YouTube. We chose a parcour clip with movements of varying complexity in different outdoor scenes. Qualitative results of these, including limitations, are shown in the supplemental video.

For quantitative evaluation of estimated height $\hat{h}$ and conversion factor $\hat{q}$ we compute absolute and signed errors to report:
\begin{itemize}
	\item \textbf{accuracy} across trials as the Mean Absolute Error (MAE) between predicted and estimated quantities.
	\item \textbf{bias} across trials as the Mean signed Error (ME).
\end{itemize}
Both measures are computed for absolute and relative errors since relative errors are easier to grasp for the small values of $q$. For consistency analysis, we report the Standard Deviation (SD) for absolute and signed errors using the $\pm$ notation. 

The ground truth ratio $q$ is computed from a reference object of known height and its height in the images.

\subsection{Estimating a Rigid Object's Height}

\begin{figure}[t]
	\begin{center}
		\includegraphics[width=7.5cm]{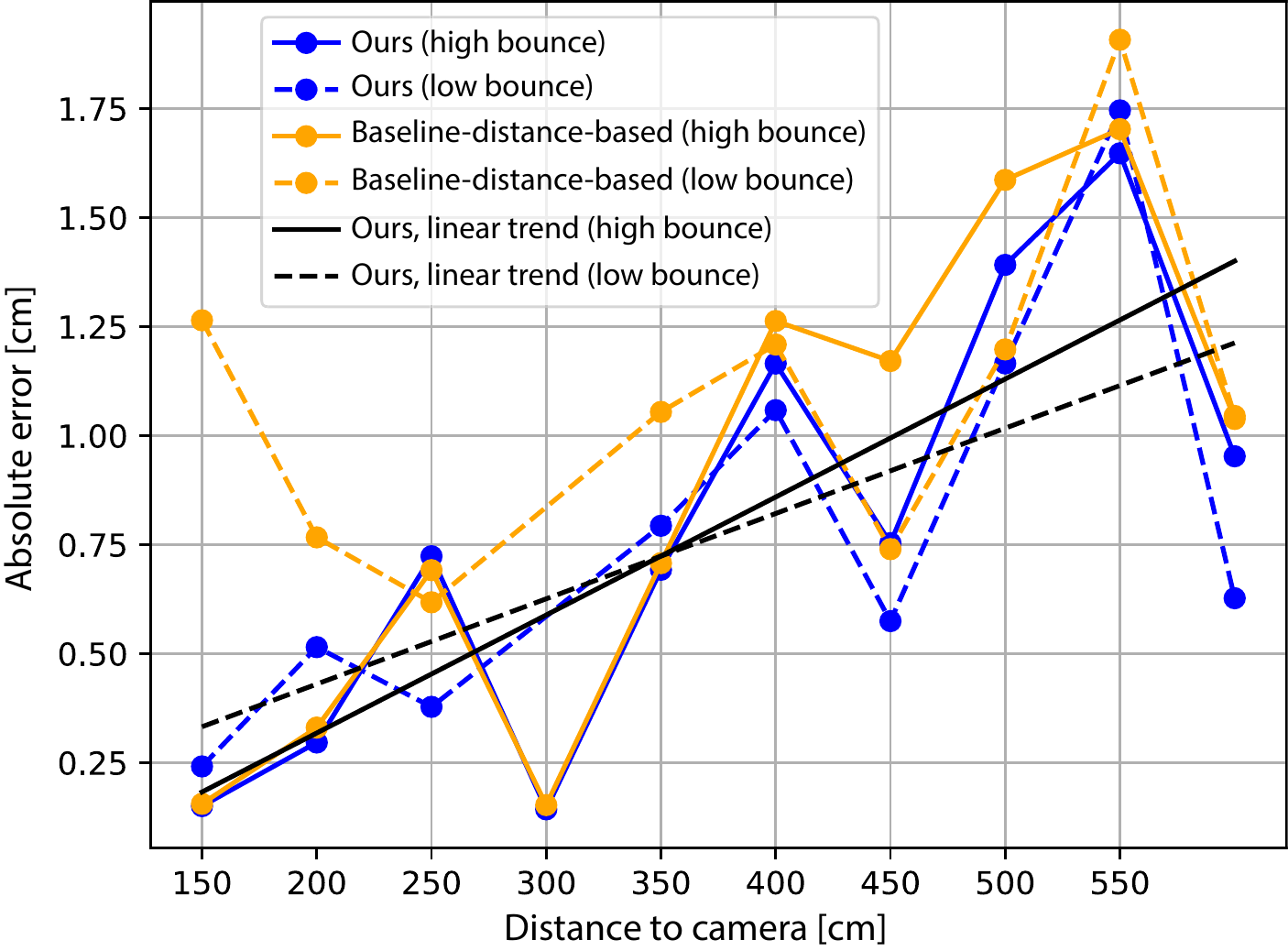}
		\caption{\textbf{Absolute size estimation error} as a function of the distance to the camera. The error is small overall, below one centimeter for distances up to 4m and follows a linear trend with respect to the distance. The varied bounce heights do not influence the accuracy significantly.}
		\label{fig:tradeoffDistance}
	\end{center}
\end{figure}

This experiment analyzes height estimation accuracy on a tennis ball being thrown from varying heights and distances to the camera.
Figure~\ref{fig:tradeoffDistance} plots the absolute error with respect to the distance to the camera. Overall, the error is small and below one centimeter for distances up to four meters. Acceleration estimation with curve fitting, \curve{}, slightly outperforms the simpler distance-based strategy, \dist{}.
This marginal difference can be explained with the high accuracy of the ball detection, larger differences can be seen in the subsequent evaluation on persons.

To analyze the overall behavior we show the linear fit of errors from \curve{} as black lines in Figure~\ref{fig:tradeoffDistance}. The error increases with respect to the distance from the camera. This is expected, as the projected object size and motion decrease in scale proportionally to the distance while the image estimation error stays constant due to the pixel discretization. 
We analyzed two different bounce heights. Although the lower bounce is roughly 50\% smaller, we found no significant influence on the scale estimation accuracy. This can be seen on the linear fits in Figure~\ref{fig:tradeoffDistance}.

\subsection{Estimating a Person's Height}
\begin{table*}[t]
	\caption{\textbf{Error analysis per subject for on-spot jumps.} We compute the median prediction across the four jumps of each subject and report the relative errors, absolute errors, as well as the bias (ME) and accuracy (MAE) across all subjects. Estimates are accurate and, in contrast to \cite{Gunel18}, unbiased with 3.9 MAE and 0.8 ME.}
	\label{tab:errorPerSubject}
	\centering
	\resizebox{0.8\linewidth}{!}{
		\begin{tabular}{|l|c|c| c|c|c|c| c|c|c|c| c|c|}
\multicolumn{13}{l}{Detailed results using \curve{}:}\\
			\hline
			\multicolumn{3}{|c|}{\bf Subjects} & \multicolumn{4}{|c|}{\bf 4m distance} & \multicolumn{4}{|c|}{\bf 7m distance} & \multicolumn{2}{|c|}{\bf Combined 4m and 7m} \\
			ID & Height & Sex & Error & Rel. & Error & Rel. & Error & Rel. & Error & Rel. & Error & Rel.\\
			& [cm] & [f or m] & [cm] & [\%] & [cm/px] & [\%] & [cm] & [\%] & [cm/px] & [\%] & [cm] & [\%]\\
\hline
S1 & 163.0 & m& -2.4 & -1.5 & -0.015 & -3.7 & 3.0   & 1.8  & 0.035  & 4.8  & 1.3  & 0.8  \\
S2 & 178.0 & m& -1.0 & -0.6 & 0.000  & 0.0  & -3.3  & -1.9 & 0.030  & 4.1  & -2.4 & -1.4 \\
S3  & 175.0 & m& 0.6  & 0.3  & 0.000  & 0.0  & 2.6   & 1.5  & -0.025 & -3.4 & 1.3  & 0.7  \\
S4  & 159.0 & f& -7.4 & -4.7 & -0.030 & -7.3 & 2.3   & 1.5  & 0.005  & 0.7  & -2.6 & -1.6 \\
S5 & 175.0 & m& 0.3  & 0.2  & 0.010  & 2.4  & -9.1  & -5.2 & -0.040 & -5.5 & -2.8 & -1.6 \\
S6  & 157.0 & f& 0.8  & 0.5  & 0.000  & 0.0  & 1.1   & 0.7  & 0.005  & 0.7  & 1.1  & 0.7  \\
S7  & 188.0 & m& -5.4 & -2.9 & -0.005 & -1.2 & -2.2  & -1.2 & 0.025  & 3.4  & -4.7 & -2.5 \\
S8  & 183.0 & m& -3.2 & -1.8 & 0.010  & 2.4  & -11.2 & -6.1 & -0.015 & -2.1 & -6.1 & -3.3 \\
S9  & 163.0 & f& 23.4 & 14.4 & 0.050  & 12.2 & -0.4  & -0.3 & -0.010 & -1.4 & 13.2 & 8.1  \\
S10  & 170.0 & f& 4.9  & 2.9  & 0.000  & 0.0  & 10.1  & 6.0  & 0.015  & 2.1  & 7.9  & 4.7  \\
S11  & 173.0 & m& -3.9 & -2.3 & -0.025 & -6.1 & 2.6   & 1.5  & 0.025  & 3.4  & 0.0  & 0.0  \\			
S12  & 173.0 & m& 0.2  & 0.1  & 0.005  & 1.2  & 7.9   & 4.5  & 0.025  & 3.4  & 3.4  & 2.0  \\
			\hline
			\multicolumn{13}{l}{Bias (ME$\pm$STD) across subjects:}\\
\hline
\multicolumn{3}{|l|}{\curve{}} & 0.6$\pm$7.9 & 0.4$\pm$4.8 & 0.00$\pm$0.02 & 0$\pm$4.9 & 0.3$\pm$6.1 & 0.2$\pm$3.5 & 0.01$\pm$0.02 & 0.9$\pm$3.3 & 0.8$\pm$5.4 & 0.6$\pm$3.2 \\

			\multicolumn{3}{|l|}{\dist{}} & -7$\pm$21.3 & 4.3$\pm$12.5 & 0.02$\pm$0.05 & 4$\pm$13.1 & -5.7$\pm$21.1 & 3.3$\pm$12 & 0.02$\pm$0.09 & 2.2$\pm$12.2 & -7.7$\pm$16.5 & 4.5$\pm$9.5 \\
			\multicolumn{3}{|l|}{\textbf{Baseline-population-mean}} & n/a& n/a& n/a& n/a& n/a& n/a& n/a& n/a& -2.5$\pm$9.5 & n/a \\
			\multicolumn{3}{|l|}{\textbf{Gunel et al.~\cite{Gunel18}}} & 3.7$\pm$6.8 &	2.3$\pm$4 & n/a & n/a & 3.6$\pm$6.9 & 2.3$\pm$4.1 & n/a & n/a & 3.7$\pm$6.7 & 2.3$\pm$4\\
\hline
\multicolumn{13}{l}{Accuracy (MAE$\pm$STD) across subjects:}\\
\hline
\multicolumn{3}{|l|}{\curve{}} & 4.5$\pm$6.4 & 2.7$\pm$3.9 & 0.01$\pm$0.02 & 3$\pm$3.8 & 4.6$\pm$3.8 & 2.7$\pm$2.1 & 0.02$\pm$0.01 & 2.9$\pm$1.6 & 3.9$\pm$3.7 & 2.3$\pm$3.2 \\
			\multicolumn{3}{|l|}{\dist{}} & 18.3$\pm$11.9 & 10.8$\pm$7 & 0.05$\pm$0.03 & 11.7$\pm$6.4 & 13.6$\pm$16.7 & 7.8$\pm$9.5 & 0.05$\pm$0.07 & 7.2$\pm$9.9 & 15.5$\pm$8.8 & 9$\pm$5\\
			\multicolumn{3}{|l|}{\textbf{Baseline-population-mean}} & n/a& n/a& n/a& n/a& n/a& n/a& n/a& n/a& 8.1$\pm$5.0& n/a\\
			\multicolumn{3}{|l|}{\textbf{Gunel et al.~\cite{Gunel18}}} & 6.5$\pm$4 & 3.8$\pm$2.5 & n/a & n/a & 6.5$\pm$3.9 & 3.8$\pm$2.5 & n/a & n/a & 6.5$\pm$3.9 & 3.8$\pm$2.4\\
			\hline
		\end{tabular}
	}
\end{table*}
\begin{table*}[t]
	\caption{\textbf{Error analysis per on-spot jump type.} We compute the accuracy (MAE) and bias (ME) across all 12 subjects. Estimation accuracy is roughly the same for all jump heights, with outliers in each class leading to high mean errors and standard deviations. This highlights the importance of taking multiple measurements.}
	\label{tab:errorPerJump}
	\centering
	\resizebox{0.8\linewidth}{!}{
	\begin{tabular}{|c| c|c|c|c| c|c|c|c|}
\hline
 & \multicolumn{4}{|c|}{\textbf{4m distance}} & \multicolumn{4}{|c|}{\textbf{7m distance}} \\
& \multicolumn{2}{|c|}{\bf Error in h} & \multicolumn{2}{|c|}{\bf Relative error in q} & \multicolumn{2}{|c|}{\bf Error in h} & \multicolumn{2}{|c|}{\bf Relative error in q} \\
& Bias [cm] & Accuracy [cm] & Bias [\%] & Accuracy [\%] & Bias [cm] & Accuracy [cm] & Bias  [\%] & Accuracy [\%] \\
& (ME$\pm$STD)& (MAE$\pm$STD)& (ME$\pm$STD)& (MAE$\pm$STD)& (ME$\pm$STD) & (MAE$\pm$STD) & (ME$\pm$STD) & (MAE$\pm$STD) \\
\hline
\textbf{J1 (low)} & -0.3$\pm$18.5 & 13.1$\pm$12.4 & 0$\pm$11.2   & 8$\pm$7.5   & -14.8$\pm$37.3 & 20.7$\pm$34  & -7.2$\pm$21 & 11.1$\pm$19 \\
\textbf{J2 (high)} & -0.6$\pm$12.4 & 8.0$\pm$9.1   & -0.1$\pm$8.1 & 5.8$\pm$5.4 & 1.6$\pm$7.9    & 6.5$\pm$4.2  & 2$\pm$4.8   & 4.4$\pm$2.4 \\
\textbf{J3 (jack)} & 1.1$\pm$8     & 5.6$\pm$5.6   & 1.1$\pm$5.1  & 3.5$\pm$3.8 & 0.2$\pm$7      & 5.3$\pm$4.3  & 1.3$\pm$4.3 & 3.1$\pm$3.1 \\
\textbf{J4 (funny)} & 0.4$\pm$11.9  & 9.4$\pm$6.8   & 0.8$\pm$6.8  & 5.2$\pm$4.2 & 4$\pm$14       & 11.6$\pm$8.1 & 3.6$\pm$8.1 & 7.7$\pm$3.9 \\
\hline
\end{tabular}
}
\end{table*}

Qualitative results and exemplar videos from the new \persons{} dataset are shown in the supplemental video. Here we analyze the accuracy quantitatively. All videos are processed with the described COM computation and automatic flight-phase detection using per-frame AlphaPose estimates.

\paragraph{Accuracy analysis.}
The introduced error metrics are evaluated in Table~\ref{tab:errorPerSubject}. The predictions from all four on-spot jumps are accumulated taking the median, independently for each of the 12 subjects.
The overall height estimation accuracy of 3.9 cm (MAE) is quite good given the large distance of 4 to 7 meters.
Notably is also the low bias of $0.8$ cm. 
Accuracies vary across subjects, but without apparent correlation to subject gender and height. For instance, S4 and S9 have the same height (159 and 163 cm) and gender but largely different errors. Moreover, the largest errors are distributed across all heights, e.g., S7 (188 cm) and S9 (163 cm) have both an error above 4.5 cm.
As for the tennis ball experiment, results are slightly more accurate and have lower standard deviation for jumps closer to the camera. 

Table~\ref{tab:errorPerJump} sheds light on the performance in terms of jump complexity. Low jumps have the highest error, presumably due to their short duration and resulting low numbers of samples. High and simple jumps are most accurate, followed by the symmetric jumping jack. In general, occasional AlphaPose failures occur during all jump types, resulting in relatively high MAE and SD. However, taking the median prediction across all jumps, including the difficult funny poses, increases prediction accuracy and SD significantly, as analyzed before.

\begin{figure}[t]
	\begin{center}
		\includegraphics[width=7.5cm]{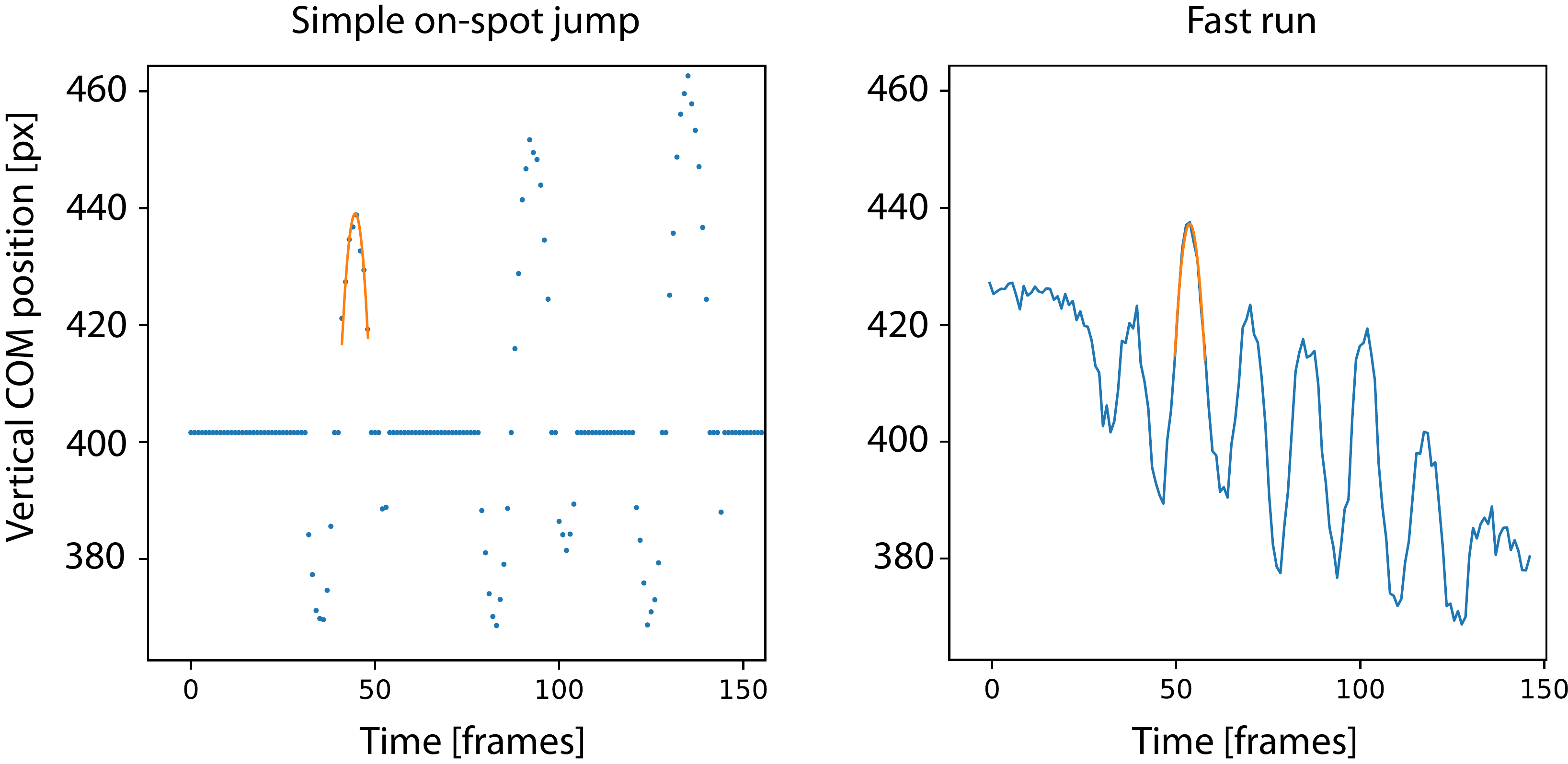}
		\caption{\textbf{COM trajectory comparison,} for 3 on-spot jumps (left) and 6 running steps (right). Running has shorter flight phases and contains temporal jitter around the peaks.}
		\label{fig:trajectoryComparison}
	\end{center}
\end{figure}

Lateral jumping and running motions are more difficult to capture due to the occluded body side. Therefore, AlphaPose exhibits many false joint detections, leading to temporal jitter in the derived COM trajectories. Table~\ref{tab:errorRunning} reports the accuracy and bias across all subjects and independently for slow run, fast run, and jumping. While Jumping is still acceptable with a MAE of 6.6 cm, the reconstructions of runs are noisy. 
Part of the error comes from the fact that flight phases during running are very short and thereby provide little data points. The difference between complex and simple jumps can be seen on the example trajectory in Figure \ref{fig:trajectoryComparison}.
This problem could be mitigated by recording with frame rates larger than the 30 Hz used here.
Another reason could be the athletic articulation leading to errors in the COM computation. 
We discuss alternative counter measures in the subsequent limitation and future work section.

\begin{table*}[]
	\caption{\textbf{Error analysis per subject for lateral motion.} Dynamic running and jumping motions lateral to the camera are hard to capture due to self-occlusions of half of the body, leading to unreliable COM trajectories and increased errors compared to the tested frontal jumps. Still, jumping is adequate with 6.6 cm MAE.}
	\label{tab:errorRunning}
	\centering
	\resizebox{\linewidth}{!}
	{
		\begin{tabular}{|c| c|c|c|c| c|c|c|c| c|c|c|c| c|c|c|c|}
			\hline
			& \multicolumn{4}{|c|}{Jumping, 4m} & \multicolumn{4}{|c|}{Slow Run, 4m} & \multicolumn{4}{|c|}{Fast Run, 4m} \\
			& \multicolumn{2}{|c|}{h} & \multicolumn{2}{|c|}{q (= 0.0041)} & \multicolumn{2}{|c|}{h} & \multicolumn{2}{l}{q (= 0.0041)} & \multicolumn{2}{|c|}{h} & \multicolumn{2}{|c|}{q (GT: = 0.0041)} \\
			& Error & Rel. & Error & Rel. & Error & Rel. & Error & Rel. & Error & Rel. & Error & Rel. \\
			& [cm] & [\%] & [cm/px] & [\%] & [cm] & [\%] & [cm/px] & [\%] & [cm] & [\%] & [cm/px] & [\%] \\
			\hline
Bias (ME+STD) & 0.4$\pm$7.5 & 0.2$\pm$4.4 & -0.01$\pm$0.02 & -1.2$\pm$4.2 & 11.7$\pm$19.7 & 6.8$\pm$11.8 & -0.02$\pm$0.04 & -5.4$\pm$9.7 & 11.4$\pm$8.5 & 6.6$\pm$5   & -0.01$\pm$0.02 & -3.2$\pm$4.5 \\
Accuracy (MAE+STD) & 6.6$\pm$3   & 3.9$\pm$1.8 & 0.01$\pm$0.01  & 3.3$\pm$2.8  & 18.9$\pm$12.2 & 11.2$\pm$7.3 & 0.03$\pm$0.03  & 7.2$\pm$8.3  & 12.3$\pm$7   & 7.1$\pm$4.1 & 0.02$\pm$0.02  & 4$\pm$3.7    \\
			\hline
		\end{tabular}
	}
\end{table*}

\parag{Quantitative comparison.}
We compare our approach to the only existing method for estimating height from monocular and uncalibrated footage \cite{Gunel18}. Since that method regresses height with a neural network from single images, we apply it to all video frames and compute error values on the median prediction across all frames, to be comparable. \cite{Gunel18} shows large errors for tall and short persons, high MAE of $6.5$ cm, and strong bias of 3.7 cm, see bottom of Table~\ref{tab:errorPerSubject}\cite{Gunel18}. This can be explained with a tendency to the mean person height due to the uncertainty of visual cues. The naive \population{}, predicting the population mean height of 168.9 m, is largely outperformed.

\parag{Ablation study.} We now turn to analyzing independent model components and alternatives.
The negligible bias of $0.5$ cm (in Table~\ref{tab:errorPerSubject}) towards overestimating height shows that the simple linear translation from ankle-nose to total height works well. This is further evidenced by the similar relative errors of $q$, which depends only on the accuracy of $a^\text{px}$, compared to that of $h$, that includes both steps. 

To validate the importance of outlier rejection, we disable the confidence-based outlier removal.
Without any rejection, strong outliers in the pose estimation distort the curve fitting, shown for one jump in Figure~\ref{fig:useRandsac}.
Using RANSAC on top helps to reduce some errors, but worsens others. In total it increases the MAE of $h$ to 7.4 cm. 

Finally, we repeat the comparison of the curve fitting, \curve{}, to the simpler \dist{} approach. Unlike for the ball experiment, \dist{} performs significantly worse, see the bottom part of Table~\ref{tab:errorPerSubject}. This is because \curve{} can tolerate moderate AlphaPose errors by fitting to the entire jump, while \dist{} is sensitive to the peak and end frame pair.

The distance dependent error due to the assumed scaled-orthographic projection model is analyzed in the supplemental document using a simulation. The document further contains an analysis on the resilience to camera rotation.

\section{Limitations and Future Work}
\label{sec:discussion}
Although our results are up to two times better than the one of G\"unel et al., there are multiple technical constraints. 
Besides requiring free-flight motion, which restricts the application scenarios, the assumption of a static camera is the strongest limitation. However, we believe that existing video stabilization \cite{Liu14b}, video panorama \cite{Agarwala05}, and SfM  techniques \cite{Schonberger16} could be applied to resolve this constraint by registering moving videos to a static frame.
In fact, our strategy could be integrated into SfM methods and provide the scale information missing in monocular reconstruction.

Detecting free-flight, is a challenge that we only touched in this study. Currently the flight and ground contact phases can only be distinguished within a hand-selected snippet that must contain flight phases. While sufficient for many tasks such as forensics and analyzing legacy videos, developing fully-automatic techniques, perhaps by analyzing foot-ground contact visually, poses an important future research direction for domains that require automation.

It is important to reinforce that our method recovers metric height, the absolute extend of an object in the direction of gravity, but no other extends. Computing scales in arbitrary directions, such as object width, would require knowledge of the 3D gravity direction and camera calibration to disentangle the effects of foreshortening and pixel~extend.

\section{Conclusion}

We have explored a new approach for estimating the height of a person jumping and running. It is applicable to monocular videos of persons and objects alike, whenever free-fall with negligible air friction is present.
We hope that this approach will enable new applications, as its limitations are complementary to those of existing monocular height estimation approaches. The precision is high, attaining up to 3.9 cm MAE on our new \persons{} dataset. In the future, we expect that advances in person keypoint detection will further improve the proposed gravity-based height estimation strategy.

\paragraph{Acknowledgement.} We thank artist \href{https://www.youtube.com/watch?v=aAtbsh2U8Uw}{\emph{Michael Dick}} for his outstanding parkour performances. This work was supported in part by the Swiss National Science Foundation.

\section{Appendix}

\begin{table}[b!]
	\caption{\textbf{Error analysis on jumping direction and distance.} Errors increase for non-fronto-parallel motions, but are negligible for medium to far distances as perspective effects diminish.}
	\label{tab:errorPerSubject}
	\centering
	\resizebox{1\linewidth}{!}{
		\begin{tabular}{l|cccc}
			AE [cm] with respect to distances $d$	& $d=4$m	& $d=7$m	& $d=15$m	& $d=30$m\\
			\hline
			Fronto-parallel ($\alpha=0^\circ$ towards cam)& 0.4 & 0.1 & 0.03 & 0.0 \\ 
			Approx. parallel ($\alpha=10^\circ$ towards cam) & 0.2 & 0.04 & 0.06 & 0.0 \\
			Diagonal jump ($\alpha=45^\circ$ towards cam) & 9.7 & 2.9 & 0.6 & 0.2\\
			Jump straight towards camera ($\alpha=90^\circ$) & 21. & 6.0 & 1.2 & 0.3\\
	\end{tabular}	}
\end{table}
\paragraph{Upright camera assumption and fronto-parallel motion.}
An upright camera was assumed because it is so in 99\% of all videos, which eases flight detection. Nevertheless, our algorithm (Sec.~3.1 main document) applies to situations with rotated camera position, where the direction of gravity deviates from the camera up direction. This is evident in the supplemental video. To make sure, we added additional jumping sequence with camera roll to our dataset.
Using the max-min flight detection, our algorithm yields {4.2, 5.7, 0.4, and 5.7} cm AE for 0, 15, 30, and 45 deg. roll. This is comparable to the 4.5 cm AE obtained for the upright test at the same distance (4 m). Similar error ranges are obtained for camera yaw and pitch.
 
\parag{Depth influence on error.}
The algorithm works for a broad range of distances, as verified quantitatively in the tennis ball experiment. As in most vision algorithms, its accuracy drops with increasing distance and decreasing resolution. Other factors, for computing a person's Center Of Mass (COM), rather decreases with the distance.

To test this, we synthesized a jumping person in 3D and projected its 3D joint locations to a virtual, perspective camera. 
We set the jump length to 1m, jump height to 15cm, and person height to 1.8m. Despite root motion, the root-centered pose was kept static to remove the effect of approximating the COM from approximate body part weights. 
This setup allows us to compare the AE of our algorithm for different motion angles to the camera, independently of measurement noise.

Table \ref{tab:errorPerSubject} shows that the COM estimation generally decreases with the dirstance.

The perspective effect is an error source when non-parallel motions are captured at small distance ($d=4$~m). However, the error reduces with distance and angle, and is negligible ($<1$~cm) for $\alpha<10^\circ$ or $d>15$~m. Therefore, non-fronto-parallel motions can be used if captured at a sufficient distance, and close distances are accurate when motion is fronto-parallel. In these cases, perspective effects are negligible and the scaled-orthographic camera assumption holds.

\parag{Dominant error.} Compared to the simulation, the real experiments have an order of magnitude larger error ($>3$~cm), which shows that the dominant source of error is the joint detection and the dependent COM estimation.

\parag{Camera model.} We use scaled-orthographic projection in our derivation (Eq.~2, main document). Interestingly, equations still hold with affine projection. We verified this in the simulation described above. Height error is zero at all configurations when using affine instead of perspective projection, as is when using scaled-orthographic projection.

\parag{Image distortion influence.} 
We conducted our initial tests on undistorted images but distortion effects appeared to be negligible. We therefore report results on raw images. %

{\small
\bibliographystyle{ieee_fullname}
\bibliography{string,misc,vision,graphics,biomed,learning,optim,prob}

\begin{thebibliography}{10}\itemsep=-1pt

\bibitem{Adjeroh10}
D. Adjeroh, D. Cao, M. Piccirilli, and A. Ross.
\newblock {Predictability and Correlation in Human Metrology}.
\newblock In {\em IEEE International Workshop on Information Forensics and
  Security}, 2010.

\bibitem{Agarwala05}
A. Agarwala, K.C. Zheng, C. Pal, M. Agrawala, M. Cohen, B. Curless, D. Salesin,
  and R. Szeliski.
\newblock Panoramic video textures.
\newblock In {\em ACM Transactions on Graphics}, volume~24, pages 821--827.
  ACM, 2005.

\bibitem{Benabdelkader08}
C. BenAbdelkader and Y. Yacoob.
\newblock {Statistical Body Height Estimation from a Single Image}.
\newblock In {\em Automated Face and Gesture Recognition}, pages 1--7, 2008.

\bibitem{Clauser71}
C.E. Clauser, J.T. McConville, and J.W. Young.
\newblock {Weight, Volume, and Center of Mass Segments of the Human Body}.
\newblock {\em Journal of Occupational and Environmental Medicine}, 13(5):270,
  1971.

\bibitem{Criminisi00}
A. Criminisi, I. Reid, and A. Zisserman.
\newblock Single view metrology.
\newblock {\em International Journal of Computer Vision}, 40(2):123--148, 2000.

\bibitem{Dantcheva18}
Antitza Dantcheva, Francois Bremond, and Piotr Bilinski.
\newblock Show me your face and i will tell you your height, weight and body
  mass index.
\newblock pages 3555--3560, 08 2018.

\bibitem{Dey14}
R. Dey, M. Nangia, W. Ross, W. Keith, and Y. Liu.
\newblock {Estimating Heights from Photo Collections: A Data-Driven Approach}.
\newblock In {\em ACM conference on Online social network}, 2014.

\bibitem{Fang17b}
H.-S. Fang, S. Xie, Y.-W. Tai, and C. Lu.
\newblock {RMPE}: Regional multi-person pose estimation.
\newblock In {\em ICCV}, 2017.

\bibitem{Fischler81}
M.A Fischler and R.C. Bolles.
\newblock {Random Sample Consensus: A Paradigm for Model Fitting with
  Applications to Image Analysis and Automated Cartography}.
\newblock {\em Communications ACM}, 24(6):381--395, 1981.

\bibitem{Georgiev13}
T. Georgiev, , Z. Yu, A. Lumsdaine, and S. Goma.
\newblock {Lytro Camera Technology: Theory, Algorithms, Performance Analysis}.
\newblock In {\em Multimedia Content and Mobile Devices}, 2013.

\bibitem{Gordon89}
C.C. Gordon, T.~Churchil land C.E.~Clauser, B. Bradtmiller, and J.T.
  McConville.
\newblock {Anthropometric Survey of US Army Personnel: Methods and Summary
  Statistics 1988}.
\newblock Technical report, Anthropology Research Project Inc Yellow Springs
  OH, 1989.

\bibitem{Guan09}
Y. Guan.
\newblock {Unsupervised Human Height Estimation from a Single Image}.
\newblock {\em Journal of Biomedical Science and Engineering}, 2009.

\bibitem{Gunel18}
S. G{\"{u}}nel, H. Rhodin, and P. Fua.
\newblock {What Face and Body Shapes Can Tell About Height}.
\newblock In {\em arXiv preprint arXiv:1805.10355}, 2018.

\bibitem{Hartley00}
R. Hartley and A. Zisserman.
\newblock {\em {Multiple View Geometry in Computer Vision}}.
\newblock Cambridge University Press, 2000.

\bibitem{Kato98}
K. Kato and A. Higashiyama.
\newblock {Estimation of Height for Persons in Pictures}.
\newblock In {\em Perception {\&} psychophysics}, 1998.

\bibitem{kim98}
T. Kim, Y. Seo, and K.-S. Hong.
\newblock Physics-based 3d position analysis of a soccer ball from monocular
  image sequences.
\newblock In {\em International Conference on Computer Vision}, pages 721--726.
  IEEE, 1998.

\bibitem{Kumar11}
A. Kumar, P.S. Chavan, V.K. Sharatchandra, S. David, P. Kelly, and N.E.
  O'Connor.
\newblock 3d estimation and visualization of motion in a multicamera network
  for sports.
\newblock In {\em Irish Machine Vision and Image Processing Conference}, pages
  15--19. IEEE, 2011.

\bibitem{Li11d}
S. Li, V.H. Nguyen, M. Ma, C. Jin, T.D. Do, and H. Kim.
\newblock {A Simplified Nonlinear Regression Method for Human Height Estimation
  in Video Surveillance}.
\newblock In {\em EURASIP Journal on Image and Video Processing}, 2011.

\bibitem{Liu14b}
S. Liu, L. Yuan, P. Tan, and J. Sun.
\newblock Steadyflow: Spatially smooth optical flow for video stabilization.
\newblock In {\em Conference on Computer Vision and Pattern Recognition}, pages
  4209--4216, 2014.

\bibitem{Ljungberg08}
J. Ljungberg and J. S{\"{o}}nnerstam.
\newblock {Estimation of Human Height from Surveillance Camera Footage -A
  Reliability Study}.
\newblock Master's thesis, KTH, 2008.

\bibitem{Mather96}
G. Mather.
\newblock {Image Blur as a Pictorial Depth Cue}.
\newblock In {\em Proc. R. Soc. Lond. B}, 1996.

\bibitem{Newton1729}
Isaac Newton.
\newblock {\em The Mathematical Principles of Natural Philosophy}.
\newblock 1729.

\bibitem{Ohno00}
Y. Ohno, J. Miura, and Y. Shirai.
\newblock {Tracking Players and Estimation of the 3D Position of a Ball in
  Soccer Games}.
\newblock In {\em International Conference on Pattern Recognition}, 2000.

\bibitem{Ribnick09}
E. Ribnick, S. Atev, and N.P. Papanikolopoulos.
\newblock Estimating 3d positions and velocities of projectiles from monocular
  views.
\newblock {\em IEEE Transactions on Pattern Analysis and Machine Intelligence},
  31(5):938--944, 2009.

\bibitem{Rogez18}
G. Rogez, P. Weinzaepfel, and C. Schmid.
\newblock {Lcr-Net++: Multi-Person 2D and 3D Pose Detection in Natural Images}.
\newblock In {\em arXiv preprint arXiv:1803.00455}, 2018.

\bibitem{Schonberger16}
J.L. Schonberger and J.-M. Frahm.
\newblock Structure-from-motion revisited.
\newblock In {\em Conference on Computer Vision and Pattern Recognition}, pages
  4104--4113, 2016.

\bibitem{Shi15b}
J. Shi, X. Tao, L. Xu, and J. Jia.
\newblock Break ames room illusion: depth from general single images.
\newblock {\em ACM Transactions on Graphics}, 34(6):225, 2015.

\bibitem{Shiang99}
T.Y. Shiang.
\newblock {A Statistical Approach to Data Analysis and 3D Geometric Description
  of the Human Head and Face}.
\newblock In {\em Proceedings of the National Science Council, Republic of
  China. Part B, Life sciences}, 1999.

\bibitem{Skold15}
J. Sk{\"o}ld.
\newblock Estimating 3d-trajectories from monocular video sequences, 2015.

\bibitem{Vester12}
J. Vester.
\newblock {Estimating the Height of an Unknown Object in a 2D Image}.
\newblock Master's thesis, KTH, 2012.

\bibitem{Zhou16d}
X. Zhou, P. Jiang, X. Zhang, B. Zhang, and F. Wang.
\newblock {The Measurement of Human Height Based on Coordinate Transformation}.
\newblock In {\em ICIC}. 2016.

\end{thebibliography}
}

\end{document}